\documentclass[11pt]{article}
\usepackage[T1]{fontenc}
\usepackage[utf8]{inputenc}
\usepackage{lmodern}
\usepackage[margin=1in]{geometry}
\usepackage{microtype}
\usepackage{graphicx}
\usepackage{booktabs}
\usepackage{multirow}
\usepackage{array}
\usepackage{xcolor}
\usepackage{caption}
\usepackage[section]{placeins}
\usepackage[numbers,sort&compress]{natbib}
\usepackage[hidelinks]{hyperref}
\usepackage{xurl}

\newcommand{\nImages}{1{,}801}
\newcommand{\nRead}{1{,}799}
\newcommand{\nDials}{5{,}385}
\newcommand{\nPilImages}{152}

\newcommand{\nMagImages}{798}

\newcommand{\agreeMag}{96.0}

\newcommand{\nHfImages}{845}

\newcommand{\agreeHf}{97.3}

\newcommand{\nMajorTotal}{164}
\newcommand{\nAuthorTotal}{20}

\newcommand{\magAAllN}{276}
\newcommand{\magAAllImg}{67}
\newcommand{\magAAllDial}{72}

\newcommand{\magAOldImg}{62}
\newcommand{\magAOldDial}{69}

\newcommand{\magACurImg}{69}
\newcommand{\magACurDial}{74}

\newcommand{\magAN}{180}

\newcommand{\magAAll}{69}
\newcommand{\magADial}{74}

\newcommand{\magBN}{180}
\newcommand{\magBOk}{34}

\newcommand{\magBAll}{21}
\newcommand{\magBDial}{25}
\newcommand{\magBDialOk}{43}
\newcommand{\magCN}{180}
\newcommand{\magCOk}{30}

\newcommand{\magCAll}{18}
\newcommand{\magCDial}{22}
\newcommand{\magCDialOk}{46}
\newcommand{\magDN}{162}
\newcommand{\magDOk}{75}

\newcommand{\magDAll}{2}
\newcommand{\magDDial}{5}

\newcommand{\magUnits}{162}
\newcommand{\magCB}{$-$4}
\newcommand{\magCBLo}{$-$10}
\newcommand{\magCBHi}{+1}
\newcommand{\magDB}{+37}
\newcommand{\magDBLo}{+28}
\newcommand{\magDBHi}{+45}
\newcommand{\magDC}{+41}
\newcommand{\magDCLo}{+33}
\newcommand{\magDCHi}{+49}

\newcommand{\hfAAll}{62}

\newcommand{\hfBOk}{54}

\newcommand{\hfCOk}{50}

\newcommand{\hfDOk}{81}

\newcommand{\hfUnits}{197}
\newcommand{\hfCB}{$-$4}
\newcommand{\hfCBLo}{$-$10}
\newcommand{\hfCBHi}{+2}
\newcommand{\hfDB}{+22}
\newcommand{\hfDBLo}{+15}
\newcommand{\hfDBHi}{+29}
\newcommand{\hfDC}{+26}
\newcommand{\hfDCLo}{+19}
\newcommand{\hfDCHi}{+33}

\newcommand{\mtwAAll}{61}

\newcommand{\mtwBOk}{52}

\newcommand{\mtwCOk}{45}

\newcommand{\mtwDOk}{80}

\newcommand{\mtwUnits}{99}
\newcommand{\mtwCB}{$-$7}
\newcommand{\mtwCBLo}{$-$15}
\newcommand{\mtwCBHi}{+1}
\newcommand{\mtwDB}{+23}
\newcommand{\mtwDBLo}{+13}
\newcommand{\mtwDBHi}{+34}
\newcommand{\mtwDC}{+30}
\newcommand{\mtwDCLo}{+20}
\newcommand{\mtwDCHi}{+40}
\newcommand{\magPairModels}{18}
\newcommand{\magPairB}{38}
\newcommand{\magPairC}{33}
\newcommand{\magPairD}{75}
\newcommand{\magClCBLo}{$-$10}
\newcommand{\magClCBHi}{+1}
\newcommand{\magClDBLo}{+21}
\newcommand{\magClDBHi}{+54}

\newcommand{\magStrictB}{19}
\newcommand{\magStrictC}{17}
\newcommand{\magStrictD}{69}
\newcommand{\hfPairModels}{11}
\newcommand{\hfPairB}{59}
\newcommand{\hfPairC}{55}
\newcommand{\hfPairD}{81}

\newcommand{\hfClDBLo}{+8}
\newcommand{\hfClDBHi}{+39}

\newcommand{\hfStrictB}{25}
\newcommand{\hfStrictC}{31}
\newcommand{\hfStrictD}{75}
\newcommand{\hfPixImages}{56}
\newcommand{\hfNoPixCB}{$-$6}
\newcommand{\hfNoPixCBLo}{$-$12}
\newcommand{\hfNoPixCBHi}{+1}
\newcommand{\hfNoPixDB}{+20}
\newcommand{\hfNoPixDBLo}{+13}
\newcommand{\hfNoPixDBHi}{+27}

\newcommand{\hfNoPixUnits}{164}
\newcommand{\magAModelAvg}{65}
\newcommand{\pilOk}{18}

\newcommand{\pilAll}{41}

\newcommand{\pilNineOk}{5}
\newcommand{\pilZeroOk}{31}

\newcommand{\hfANbPro}{54}
\newcommand{\hfANbProOf}{54}
\newcommand{\hfAFlux}{53}
\newcommand{\hfAFluxOf}{54}
\newcommand{\hfAIdeo}{2}

\newcommand{\hfASun}{8}
\newcommand{\hfASunOf}{54}
\newcommand{\magBSwap}{5}

\newcommand{\magCSwap}{8}

\newcommand{\magBWatch}{28}
\newcommand{\magBWall}{42}
\newcommand{\magBAlarm}{32}
\newcommand{\magDWatch}{76}
\newcommand{\magDWall}{78}
\newcommand{\magDAlarm}{70}

\title{It's Always 10:10:\\Reference Images Break a Bias That Prompts Only Dent}
\author{Luca Cazzaniga\\
Independent Researcher, AI-Assisted Visual Production\\
\texttt{luca@lucacazzaniga.com}}
\date{October 2026}

\begin{document}
\maketitle

\begin{abstract}
Text-to-image models appear to reproduce the habits of the photographs they learned from. Analog clocks are an extreme case: in advertising, watches almost always show 10:10, and generated clocks return to 10:10 even when another time is requested. We measure this bias and test three ways of overcoming it on 52 models available on the Magnific platform, with a replication on Higgsfield. Every image shows three identical clocks that must show 2:35, 6:50 and 11:20. The description of the object is fixed and only the request about the time changes: no time (A), the time in digits (B), the hand positions described by construction relative to the dial numerals (C), or the same description plus a drawn reference dial (D). Two AI readers read all \nRead{} images blind from coded copies, with a third reader and the author settling disagreements (dial-level agreement \agreeMag\% and \agreeHf\%). With no time requested, \magAAllImg\% of the images have all three clocks at 10:10. On the 20 current models, all three clocks are correct in \magBOk\% of the images with digits, \magCOk\% with the hands described in words and \magDOk\% with the reference dial (D$-$B: \magDB{} points, 95\% CI \magDBLo{} to \magDBHi{}); we found no evidence that describing the hands in words beats the digits (C$-$B: \magCB{} points, CI \magCBLo{} to \magCBHi{}). The replication on the 12 models shared by both platforms gives the same picture (B \hfBOk\%, C \hfCOk\%, D \hfDOk\%). Writing the time reduces the bias but leaves two thirds of the images of the 20 current models with at least one wrong clock; adding a drawn reference raises full accuracy to three quarters and almost eliminates images entirely at 10:10. We release all images, prompts, raw readings and a script that recomputes every result.
\end{abstract}

\begin{figure}[!h]
\centering
\includegraphics[width=\linewidth]{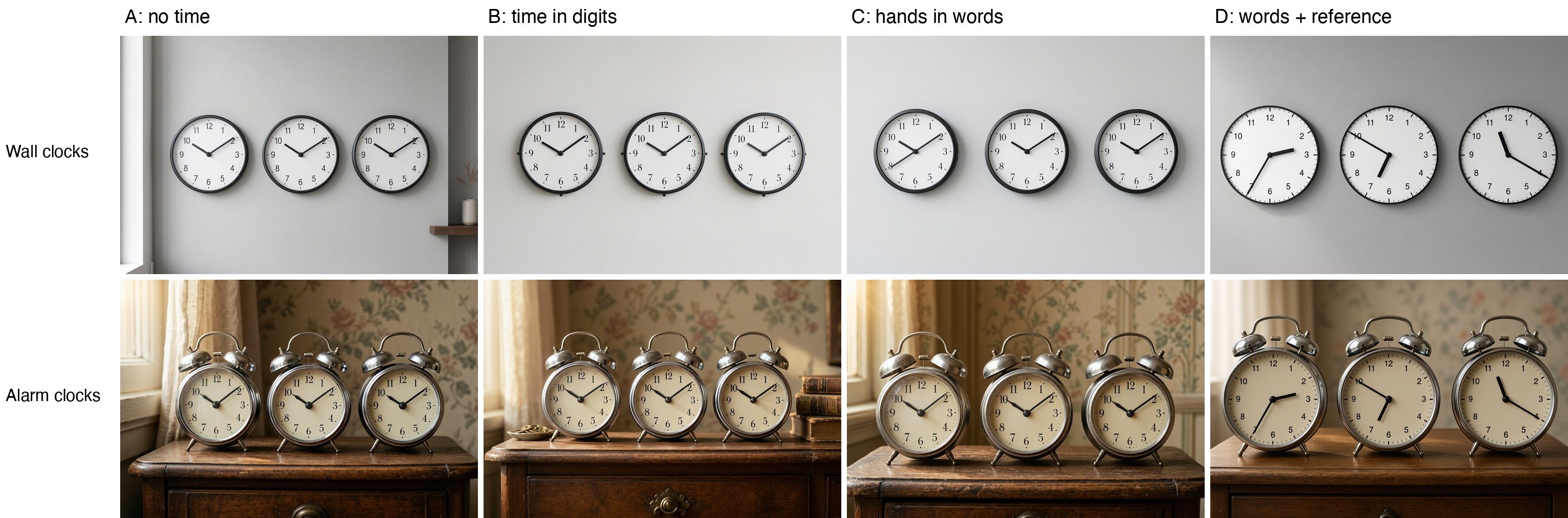}
\caption{The same model (MAI Image 2.5 on Magnific, round 1) under the four conditions. The target times are 2:35, 6:50 and 11:20 from left to right. With no time (A), with the time in digits (B) and with the hand positions described in words (C) every clock shows 10:10. With the description plus a drawn reference dial (D) all three clocks in both images show the requested time.}
\label{fig:teaser}
\end{figure}

\section{Introduction}
Ask an image generator for a watch showing a quarter to three and, more often than not, it will show ten past ten. The pose is not an accident of one model. Watch and clock advertising has used 10:10 for decades, a pose that experiments associate with a more positive emotional response to the product \citep{karim2017tenpast}. The photographs that fill training sets presumably inherit the convention, and the generators inherit it from them. The phenomenon is well known among practitioners and has been described in the press \citep{harris2025dcw,pot2026popsci}.

Clocks are a convenient test case for a more general question. Generators repeat the defaults of their training data, and sometimes no written instruction moves them: the full wine glass is another well-known example \citep{pyatko2025wine}. A clock has a property that most defaults lack: an objective ground truth. Two hand angles say whether the instruction was followed, so the bias can be measured without aesthetic judgment.

We ask four questions. (1) How strong is the 10:10 default when no time is requested? (2) Does writing the time in digits fix it? (3) Does describing the hand geometry in words, relative to the dial numerals, help more than the digits? SCHEMA \citep{cazzaniga2026schema} is a structured prompting method that writes the request as labelled sections (subject, design, setting, composition, camera, lighting, style, constraints) and recommends describing spatial requirements by construction, relative to visible landmarks, rather than by name or number; condition C applies that recommendation to the hands. (4) Does attaching a drawn reference dial with the right angles help more than the wordings we tested?

\paragraph{Contributions.}
\begin{itemize}
\item A controlled comparison of four prompting conditions on 52 text-to-image models available on one platform (Magnific), with three rounds on the 20 current models and a replication of the 12 models also available on a second platform (Higgsfield) with six rounds: \nImages{} images generated, \nRead{} read, three clocks each.
\item A blind reading protocol with two independent AI readers, a third reader on disagreements and human arbitration of the remaining cases; the two readers assign the same class to \agreeMag\% of the dials in the main study and \agreeHf\% in the replication.
\item The finding that the time written in digits reduces the bias but leaves most images of the 20 current models with at least one wrong clock, that we found no evidence that a precise description of the hands in words does better than the digits, and that adding a drawn reference dial raises the share of fully correct images by \magDB{} points on the main study and \hfDB{} points on the shared models.
\item An open dataset with every image, prompt, raw reading and verdict, and a script that recomputes all results from the raw readings \citep{cazzaniga2026dataset}.
\end{itemize}

\section{Related work}
\paragraph{The 10:10 bias.} The preference of advertising for 10:10 was studied experimentally by \citet{karim2017tenpast}, who found that the pose increases the positive emotional response to the product. In generated images the default has been documented in the press with examples from several tools \citep{harris2025dcw,pot2026popsci}. \citet{mohammed2026watch} propose a model-side algorithm that reduces the temporal bias of watch images generated by Stable Diffusion~1.5. Our study is complementary: we do not modify any model but measure, across 52 commercial models, what can be obtained by changing only the request.

\paragraph{Reading clocks.} Reading analog clocks is itself hard for vision models. \citet{yang2022clock} train a dedicated reader on images in the wild; ClockQA \citep{saxena2025lost} and ClockBench \citep{safar2025clockbench} show that multimodal language models often misread the time; \citet{choi2026ticktock} address in particular the confusion between hour and minute hands. OmniClock \citep{omniclock2026} evaluates both reading and generation of clocks, scoring generated images with a fixed time reader and accuracy within five minutes, the same tolerance we use for the minute hand. These results motivate our protocol: two independent readers, explicit reporting of hand positions rather than of the time alone, arbitration of disagreements, and a dedicated class for swapped hands.

\paragraph{Faithfulness benchmarks.} Text-to-image alignment is usually measured with object-level checks or question answering: GenEval \citep{ghosh2023geneval}, T2I-CompBench \citep{huang2023t2icompbench}, TIFA \citep{hu2023tifa}, DPG-Bench \citep{hu2024ella}, GenAI-Bench \citep{li2024genaibench} and VQAScore \citep{lin2024vqascore} cover counting, attributes, spatial relations and composition. Hand angles are a fine-grained spatial attribute that these benchmarks do not isolate, and the 10:10 default makes failures systematic rather than random.

\paragraph{Learned defaults.} Generators reproduce regularities and sometimes specific items of their training data, from social stereotypes \citep{luccioni2023stable} to memorised images \citep{carlini2023extracting,somepalli2022forgery}. A recurrent pose such as 10:10 is a different phenomenon from the copy of a single image, and our design does not identify its origin: it measures how hard it is to move.

\paragraph{Visual conditioning.} Images can constrain generation more directly than text: ControlNet adds structural conditions such as edges and poses \citep{zhang2023controlnet}, IP-Adapter conditions on image prompts \citep{ye2023ipadapter}, Paint by Example edits from exemplars \citep{yang2022paint}, and MultiRef evaluates generation from several visual references, with persistent failures \citep{chen2025multiref}. We test the simplest form available to an end user of a commercial platform: a single reference image attached to the request.

\paragraph{Models as judges.} Multimodal models are increasingly used to evaluate generated images, with known biases and inconsistencies \citep{ku2024viescore,chen2024mllmjudge,saxon2024ts2,kamath2025geneval2}; collaborative discussion among several judges has also been proposed \citep{qian2026collabeval}. Our two-of-three vote is a simpler arbitration that those works do not validate. We use AI readers for a narrow perceptual task, under blinding, and report their agreement and the human arbitration.

\section{Method}
\subsection{Scenes, targets and conditions}
Each image contains three identical analog clocks in a row, in one of three scenes: wristwatches in a top-down product shot (S1), wall clocks (S4) and vintage twin-bell alarm clocks (S5).\footnote{The numbering keeps the original scene list. Two further scenes, worn wristwatches (S2) and dials without numerals (S3), were dropped after a graphic test because the dials were rotated or impossible to read.} The three clocks must show 2:35, 6:50 and 11:20, three times far from 10:10 and from each other. Their order rotates over three rounds (rotation 1: 2:35, 6:50, 11:20; rotation 2: 6:50, 11:20, 2:35; rotation 3: 11:20, 2:35, 6:50), so that position on the image is not confounded with the target.

\begin{figure}[htbp]
\centering
\includegraphics[width=0.62\linewidth]{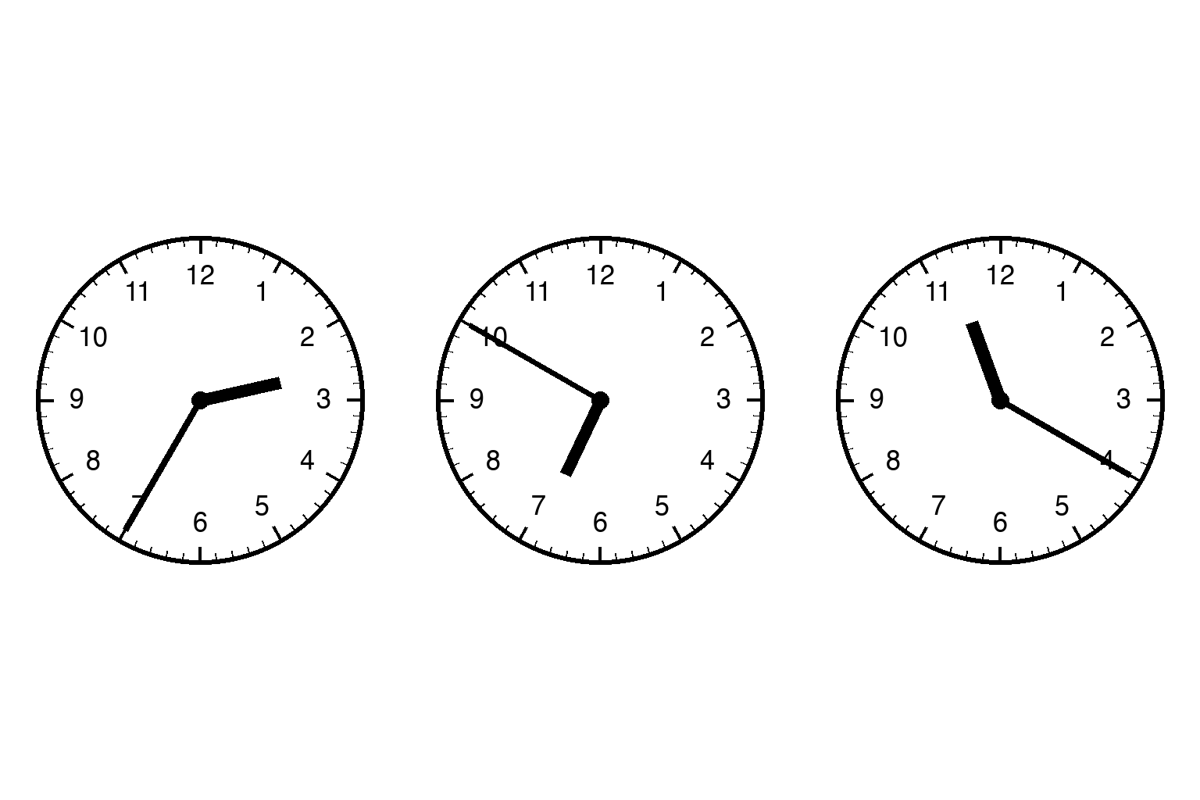}
\caption{The reference image of condition D for rotation 1 (2:35, 6:50, 11:20). It was drawn with a script, without any generative model; rotations 2 and 3 use the same drawing with the times reordered.}
\label{fig:reference}
\end{figure}

The description of the object is identical in all conditions, including the construction of the hands (a short thick hour hand reaching about half the radius, a long thin minute hand reaching the minute track, the same central pin), the scene, the light and the framing. Only the request about the time changes: one added block of lines and, in B to D, the words ``each one set as listed above'' in the constraint line (Appendix~\ref{app:prompts}; every prompt is in the dataset):
\begin{itemize}
\item \textbf{A, no time.} No time is requested: the control shows where free hands go.
\item \textbf{B, time in digits.} ``Times, left to right: 2:35, 6:50, 11:20''.
\item \textbf{C, hands in words.} Each hand is placed by construction relative to the dial numerals, without digits: for 2:35, ``the hour hand points seven twelfths of the way from the 2 to the 3; the minute hand points exactly at the 7''. Positions are never given in degrees or percentages, which models tend to print on the image.
\item \textbf{D, words plus reference.} The text of C plus a drawn reference image (Figure~\ref{fig:reference}) and one line: ``use the attached diagram solely to determine the hand angles and their left-to-right assignment. Render the clocks using the design, scene and photographic style specified here.''
\end{itemize}
The reference dials were drawn programmatically, not generated, one image per rotation. The same reference is used for all three scenes and is attached as a generic image reference, with no weight or strength parameter. Because the reference also shows a plain dial, condition D estimates the combined effect of reference conditioning, including possible changes in the appearance of the dials, not the effect of the geometric information alone.

\subsection{Models, platforms and rounds}
All images were generated on 7 October 2026 through two commercial platforms that give access to many third-party models: Magnific and Higgsfield, with paid subscriptions. Models are named as the platforms list them; the identifier used by each platform is recorded in the dataset.\footnote{On both platforms the Nano Banana names are shifted: the Magnific mode \texttt{imagen-nano-banana-2} is Nano Banana Pro and \texttt{imagen-nano-banana-2-flash} is Nano Banana~2; Higgsfield reports Nano Banana Pro jobs as ``nano\_banana\_2'' and Nano Banana~2 jobs as ``nano\_banana\_flash''.} Images were requested in 3:2 at the lowest resolution offered (1K; 1.5K or 2K for the Seedream models) and at medium quality where a choice exists; on Higgsfield, where the option is exposed, Ideogram's prompt rewriting was turned off. Mystic 2.5 Fluid has no 3:2 format and was run in 16:9 with the same prompt.

Each cell received exactly one request and its output was kept, with no selection; a request was repeated only when the server returned no image. Requests were sent in parallel batches mixing conditions and scenes; their order and identifiers are in the raw logs. Seeds are not exposed by the platforms.

\textbf{Magnific study.} Condition A was run on all 52 image models of the platform that accept a text-only request.\footnote{Runway refused requests without a reference image.} The 20 current models (Table~\ref{tab:models}), classified as such before data collection as the latest generation of each provider's line on the platform together with its variants (for example Nano Banana 2.1, Pro, 2 and 2 Lite), were run in all four conditions for three rounds, one image per model, scene and round; the other 32 models were run once in condition A (Appendix~\ref{app:old}). Two models accept only style or character references and could not run condition D (Recraft V4.1, Mystic 2.5). In total: \nMagImages{} images.

\textbf{Higgsfield replication.} The 12 current models available on both platforms were run in all four conditions for six rounds (two full rotations), with the same prompts and reference images; Recraft V4.1 again had no D. One image (Nano Banana~2 Lite, D, S5, round~3) failed three times on the server. In total: \nHfImages{} images. The two platforms are never mixed in the same cell.

\textbf{Pilot.} Before the study, condition B was run once on the 52 models with an earlier prompt version. A methodological review of that version by an independent model (GPT-6.1) led to the final prompts: the hand construction moved into the description of all conditions, contradictory constraints were removed and the precision of condition C was made uniform. The pilot is reported separately (Section~\ref{sec:pilot}).

\subsection{Blind reading}
Each image was copied as JPEG under a random six-character code into a neutrally named folder, metadata stripped and the long side reduced to at most 1600 pixels; codes were spread in shuffled batches of 13 that mix models, scenes and conditions. Readers never saw the model, the condition, the target times or the purpose of the test.

Two readers read every image, in separate sessions and without access to each other's output, with the same English instructions (released with the data): Sol (OpenAI \texttt{gpt-6.1-sol}, high reasoning effort) and Claude (Anthropic Opus~5.5 in the pilot and the Magnific study, Sonnet~5.5 in the replication). For every dial, left to right, they report where the shorter hand points (in hours, one decimal), where the longer hand points (in minutes), the number of hands, whether the two hands can be told apart, and notes. The Claude readers had a strict rule: one look per image and at most one enlarged crop, only when the hands could not be told apart; in the replication, three of the 17 Sonnet readers reported measuring hand lengths or angles on the pixels as well, for \hfPixImages{} of the 845 images; we report the replication with and without them. Sol was run through the Codex command-line tool in a read-only sandbox, one call per batch; Claude readers ran as separate Claude Code sub-agents or sessions, one per group of batches; the third reader ran through the Antigravity command-line tool with no tool other than opening the images. Dials are matched by position from left to right; when the readers count a different number of dials, the image is non-conforming.

Each reading is turned into one class from the two hand positions, never from the reported time, using circular distances on the dial and a fixed precedence: \emph{correct} (minute hand within 5 minutes and hour hand within 0.6 hours of the target; the first is the usual five-minute tolerance, the second accepts an hour hand placed on the nearest numeral instead of between numerals, a frequent drawing error, but not one placed on the preceding hour late in the hour, such as on the 6 for 6:50); \emph{10:10} (hour hand between 9.7 and 10.6, minute hand between 6 and 14); \emph{another clock's time} (the target of another clock in the same image); \emph{hands swapped} (correct once the roles are exchanged: the hour reading times five is taken as the minute position and the minute reading divided by five as the hour position); \emph{hands overlapping} (the two hands within 3 minutes of each other); \emph{other error}; and \emph{unreadable} (a position missing). Extra hands, such as second hands, are noted but do not change the class. When a reader cannot tell the hands apart, it still assigns the two roles, marks them as not distinguishable and the assignment is classified as given; when only one hand is visible, the reading either repeats the same position for both hands (classified as overlapping) or leaves one position empty (unreadable). In condition A, which has no target, everything that is not 10:10 is \emph{other time}.

When the two readers assign the same class, that class is final. Otherwise a third reader (Google Gemini~3.8 Flash) reads the image and the class shared by two of the three is final; the few dials still open were decided by the author on an annotated image. Agreement is the share of dials to which the two readers assign the same class. Of the \nDials{} dials scored, \nMajorTotal{} were settled by the majority of three readers and \nAuthorTotal{} by the author (Table~\ref{tab:counts}).

\subsection{Outcomes and statistics}
The primary outcome is image-level: \emph{all three clocks correct}. An image counts only if both readers see exactly three dials; images with more or fewer clocks stay in the denominator as failures. The secondary outcome is \emph{all three clocks at 10:10}, and we also report dial-level shares. Paired differences between conditions are computed on units \emph{model $\times$ scene $\times$ round} that have B, C and D, so models without D and the one missing image are left out of the comparison. The estimand is the difference for this fixed set of models: the main intervals come from a percentile bootstrap over units (4,000 resamples, seed 7, 95\%). As a sensitivity analysis for generalisation to similar models we also resample whole models with their units (4,000 resamples, seed 11). We report three comparisons (C$-$B, D$-$B, D$-$C) without correction for multiple comparisons. A second sensitivity analysis tightens the tolerances to 2.5 minutes and 0.3 hours and requires both readers' positions to meet them. With three rounds per cell on Magnific and six on Higgsfield, per-model results are descriptive.

\begin{table}[htbp]
\centering
\small
\begin{tabular}{lrrrr}
\toprule
Block & Generated & Read & Scored & Dials scored \\
\midrule
Magnific study & 798 & 798 & 798 & 2{,}394 \\
Higgsfield replication & 845 & 845 & 845 & 2{,}535 \\
Pilot & 156 & 156 & 152 & 456 \\
Graphic test & 2 & -- & -- & -- \\
\midrule
Total & \nImages & \nRead & 1{,}795 & \nDials \\
\bottomrule
\end{tabular}
\caption{Images and dials by block. In the pilot, four images with four clocks (seen as such by both readers) were left out; in the study and the replication, non-conforming images stay in the denominator as failures. The Higgsfield replication planned 846 images; one failed on the server.}
\label{tab:counts}
\end{table}

\section{Results}
\subsection{The default: 10:10 when no time is requested}
With no time in the prompt, \magAAllImg\% of the \magAAllN{} images of the 52 Magnific models have all three clocks at 10:10, and \magAAllDial\% of all dials are at 10:10 (\magAOldImg\% and \magAOldDial\% on the 32 older models, \magACurImg\% and \magACurDial\% on the 20 current ones; the current models contribute three rounds each, and with every model weighted equally the share of images entirely at 10:10 is \magAModelAvg\%). The default is strong but not universal. On Higgsfield, Nano Banana Pro puts \hfANbPro{} of \hfANbProOf{} dials at 10:10 and Flux.3 \hfAFlux{} of \hfAFluxOf{}, whereas Ideogram~4.5 does so for \hfAIdeo{} and GPT~2.5 Sunburst for \hfASun{} of \hfASunOf{}.

\begin{table}[htbp]
\centering
\small
\begin{tabular}{lcccc}
\toprule
Condition & Images & All three correct & All three at 10:10 & Dials at 10:10 \\
\midrule
A, no time & \magAN & (no target) & \magAAll\% & \magADial\% \\
B, time in digits & \magBN & \magBOk\% & \magBAll\% & \magBDial\% \\
C, hands in words & \magCN & \magCOk\% & \magCAll\% & \magCDial\% \\
D, words + reference & \magDN & \magDOk\% & \magDAll\% & \magDDial\% \\
\bottomrule
\end{tabular}
\caption{The 20 current models on Magnific, three rounds per model and scene. Condition D excludes the two models that accept no generic reference image.}
\label{tab:conditions}
\end{table}

\subsection{Digits, words and a reference}
\begin{figure}[htbp]
\centering
\includegraphics[width=\linewidth]{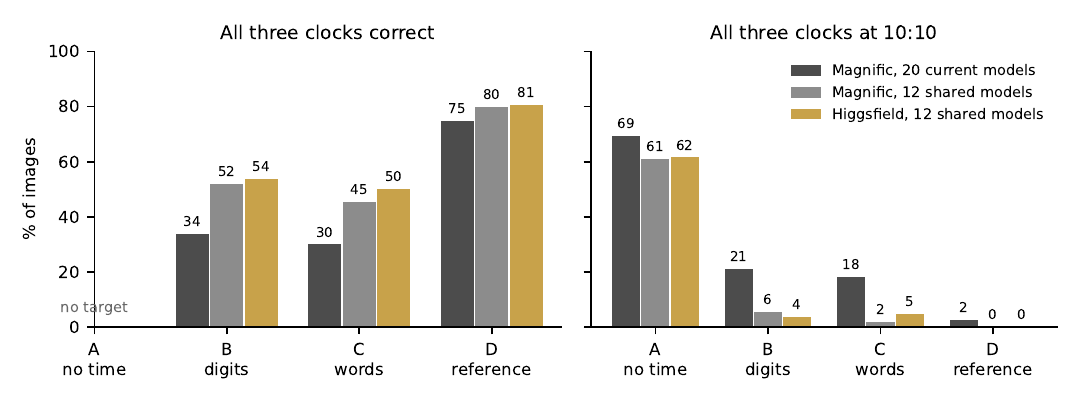}
\caption{Share of images with all three clocks correct (left) and with all three at 10:10 (right), by condition. Dark grey: the 20 current Magnific models (D on 18). Light grey and gold: the 12 models available on both platforms, on Magnific (3 rounds) and on Higgsfield (6 rounds).}
\label{fig:conditions}
\end{figure}

Table~\ref{tab:conditions} and Figure~\ref{fig:conditions} summarise the four conditions on the 20 current Magnific models. Writing the time in digits reduces the share of images entirely at 10:10 from \magAAll\% to \magBAll\%, but all three clocks are correct in only \magBOk\% of the images. We found no evidence that describing the hands in words improves on the digits: \magCOk\% of the images are fully correct, a paired difference of \magCB{} points (95\% CI \magCBLo{} to \magCBHi{}). At the level of single dials the two conditions are close (\magBDialOk\% correct dials with digits, \magCDialOk\% with words). Adding the drawn reference raises the share of fully correct images to \magDOk\%: \magDB{} points over the digits (CI \magDBLo{} to \magDBHi{}) and \magDC{} points over the words alone (CI \magDCLo{} to \magDCHi{}), on \magUnits{} paired units. On the same \magPairModels{} models that have all three conditions, the shares are \magPairB\% (B), \magPairC\% (C) and \magPairD\% (D), which is why the paired difference differs from the gap between the marginal rates. Images entirely at 10:10 almost disappear (\magDAll\%).

The conclusions survive both sensitivity analyses. Resampling whole models widens the intervals (D$-$B \magClDBLo{} to \magClDBHi{}, C$-$B \magClCBLo{} to \magClCBHi{}) without changing their sign. With the tighter tolerances, the share of fully correct images falls to \magStrictB\% with digits, \magStrictC\% with words and \magStrictD\% with the reference on Magnific (\hfStrictB\%, \hfStrictC\% and \hfStrictD\% on Higgsfield): the advantage of the reference grows, while the comparison between digits and words stays mixed.

The effect holds in every scene: with digits, \magBWatch\% of wristwatch images, \magBWall\% of wall-clock images and \magBAlarm\% of alarm-clock images are fully correct; with the reference, \magDWatch\%, \magDWall\% and \magDAlarm\%.

\begin{table}[htbp]
\centering
\small
\begin{tabular}{lccc}
\toprule
 & Magnific, 20 models & Magnific, 12 shared & Higgsfield, 12 shared \\
\midrule
Paired units & \magUnits & \mtwUnits & \hfUnits \\
C $-$ B & \magCB{} [\magCBLo, \magCBHi] & \mtwCB{} [\mtwCBLo, \mtwCBHi] & \hfCB{} [\hfCBLo, \hfCBHi] \\
D $-$ B & \magDB{} [\magDBLo, \magDBHi] & \mtwDB{} [\mtwDBLo, \mtwDBHi] & \hfDB{} [\hfDBLo, \hfDBHi] \\
D $-$ C & \magDC{} [\magDCLo, \magDCHi] & \mtwDC{} [\mtwDCLo, \mtwDCHi] & \hfDC{} [\hfDCLo, \hfDCHi] \\
\bottomrule
\end{tabular}
\caption{Paired differences in the share of images with all three clocks correct, in percentage points, with 95\% bootstrap intervals. Units are model $\times$ scene $\times$ round with conditions B, C and D.}
\label{tab:diff}
\end{table}

\subsection{Replication on a second platform}
On the 12 models available on both platforms, Higgsfield gives \hfBOk\% fully correct images with digits, \hfCOk\% with words and \hfDOk\% with the reference; the same models on Magnific give \mtwBOk\%, \mtwCOk\% and \mtwDOk\% (Figure~\ref{fig:conditions}, Table~\ref{tab:diff}). With no time, \hfAAll\% of the Higgsfield images are entirely at 10:10, against \mtwAAll\% on Magnific. The two platforms agree within a few points in every condition. Aggregate rates were similar across platforms for the shared models; the 20-model rates are lower because they include models that Higgsfield does not offer. The replication uses the same models, so it tests the platform and the generation run, not a new sample of models. On the \hfPairModels{} Higgsfield models with all three conditions the shares are \hfPairB\%, \hfPairC\% and \hfPairD\%; the gain of the reference is smaller there (\hfDB{} points; resampling models, \hfClDBLo{} to \hfClDBHi{}) because several of these models are already near the ceiling with digits. Leaving out the \hfPixImages{} images read with pixel measurements gives D$-$B \hfNoPixDB{} points (\hfNoPixDBLo{} to \hfNoPixDBHi{}) and C$-$B \hfNoPixCB{} (\hfNoPixCBLo{} to \hfNoPixCBHi{}) on \hfNoPixUnits{} units.

\subsection{Models}
\begin{figure}[htbp]
\centering
\includegraphics[width=0.55\linewidth]{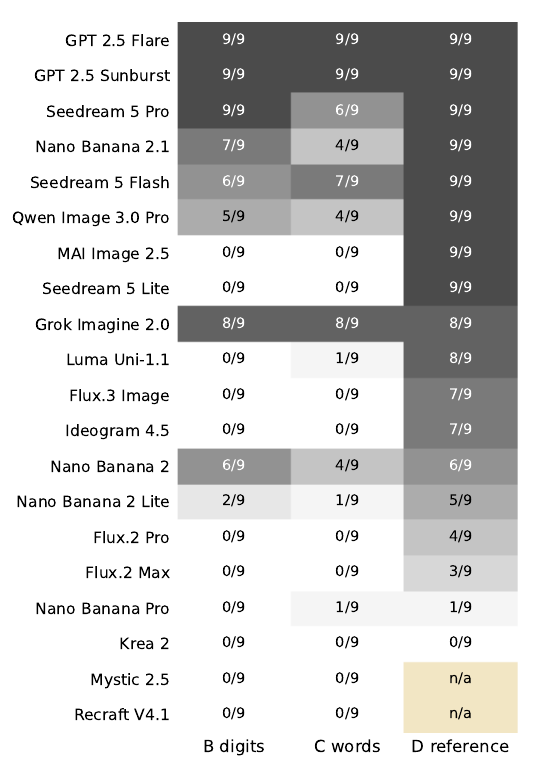}
\caption{Images with all three clocks correct out of 9, per model and condition, on Magnific. ``n/a'': the model accepts no generic reference image.}
\label{fig:models}
\end{figure}

Table~\ref{tab:models} and Figure~\ref{fig:models} show four patterns, descriptive given the few rounds per model. GPT~2.5 Flare and Sunburst draw every requested time in every condition with text alone, and Seedream~5 Pro, Seedream~5 Flash and Nano Banana~2.1 come close. A second group follows the digits poorly or not at all but draws the right time once the reference is attached: MAI Image~2.5 and Seedream~5 Lite go from 0 to 9 of 9 images, Flux.3 and Ideogram~4.5 from 0 to 7 (Ideogram reaches 17 of 18 on Higgsfield). Some models ignored the time request entirely: Flux.2 Max, Flux.2 Pro and MAI Image~2.5 put every dial at 10:10 in conditions A, B and C alike. A last group stays weak even with the reference: Krea~2, Nano Banana Pro and the two Flux.2 models.

\begin{table}[htbp]
\centering
\footnotesize
\setlength{\tabcolsep}{4pt}
\begin{tabular}{lcccc|cccc}
\toprule
 & \multicolumn{4}{c|}{Magnific (3 rounds)} & \multicolumn{4}{c}{Higgsfield (6 rounds)} \\
Model & A: 10:10 dials & B & C & D & A: 10:10 dials & B & C & D \\
\midrule
\csname @@input\endcsname tables/models.tex 
\bottomrule
\end{tabular}
\caption{Per model. Column A: dials at 10:10 with no time requested. Columns B, C, D: images with all three clocks correct. ``--'': condition not possible (no generic reference image).}
\label{tab:models}
\end{table}

\subsection{Error types}
Table~\ref{tab:errors} breaks the dials of the 20 current Magnific models into classes. With digits or words, about a quarter of the dials still go to 10:10 and about a fifth are other errors; swapped hands are slightly more frequent when the hands are described in words (\magCSwap\%) than with digits (\magBSwap\%). With the reference every class of error is less frequent.

\begin{table}[htbp]
\centering
\small
\begin{tabular}{lcccc}
\toprule
Class & A & B & C & D \\
\midrule
\csname @@input\endcsname tables/errors.tex 
\bottomrule
\end{tabular}
\caption{Share and number of dials in each class, 20 current Magnific models (D on 18 models). All dials read are included, also those of non-conforming images. Percentages are rounded and may not sum to 100.}
\label{tab:errors}
\end{table}

\subsection{Pilot: the time as text, the hands at 10:10}
\label{sec:pilot}
\begin{figure}[htbp]
\centering
\includegraphics[width=0.8\linewidth]{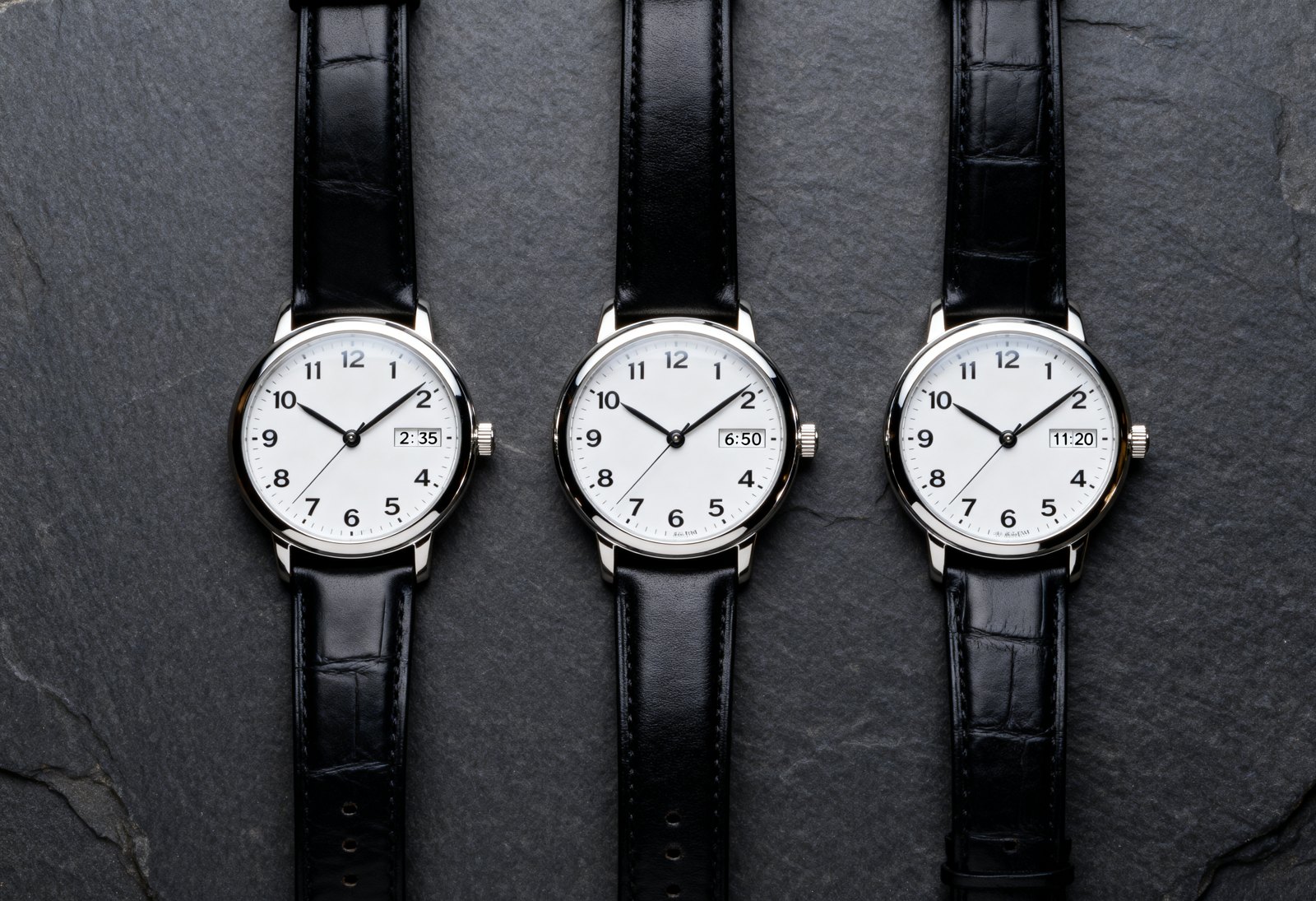}
\caption{Pilot, Seedream~4 4K, time in digits: the requested times appear in the date windows (2:35, 6:50, 11:20) while all three hands show 10:10.}
\label{fig:pilot}
\end{figure}

In the pilot (condition B, earlier prompt, one round on 52 models), all three clocks were correct in \pilOk\% of the \nPilImages{} readable images and entirely at 10:10 in \pilAll\%; \pilNineOk{} models drew every requested time and \pilZeroOk{} drew none. Twelve models put all nine dials at 10:10 despite the requested times. In six images the model wrote the requested time as text on the dial, in a date window or in place of a numeral, while the hands stayed at 10:10 (Figure~\ref{fig:pilot}): the time was understood as a string but not turned into hand positions.

\section{Discussion}
\paragraph{Showing and telling.} The prompts we tested reduce the bias but do not remove it: with the time in digits, between half and two thirds of the images still have at least one wrong clock, and on the 20 current models a quarter of the dials remain at 10:10. Spelling out the geometry, which is what a careful prompt writer would try next, gave no measurable gain over the digits. Adding a drawn reference raised the share of fully correct images to three quarters on the 20 current models and to four fifths on the shared ones. Among the models that ignored the time request, the reference turned zero correct images into nine of nine for MAI Image~2.5, while the gain was partial for the two Flux.2 models.

\paragraph{Why words did not help more.} The pilot images with the time printed in a window suggest one reading: the request is understood, but the mapping from a described configuration to drawn geometry is weak, while the prior on the pose of the hands is strong. Reference conditioning may reduce this difficulty, but our design does not identify the mechanism, and other explanations, such as the attention given to long constraint lines, remain open.

\paragraph{What the reference also transfers.} In condition D some models copied the look of the drawing as well as its angles: dials became flatter and closer to the diagram (Figure~\ref{fig:teaser}). The object is therefore not perfectly fixed in D, and part of the gain may come from a simpler dial being easier to draw, or to read, correctly. D estimates the combined effect of reference conditioning; separating the geometric information from the appearance would need further controls, such as the same diagram without hands, a reference drawn in the photographic style of the target object, or the time in digits plus a reference. We leave them to a follow-up study.

\paragraph{Implications.} For practitioners facing a strong default, our results suggest attaching a simple drawing of the intended configuration rather than rewording the request. As a practical suggestion that this study does not test, a structured description such as SCHEMA \citep{cazzaniga2026schema} can still specify the scene and help plan the reference. For evaluation, the clock is an inexpensive probe of instruction following with an objective answer, and in our data it separated the models sharply. Whether the same pattern holds for other strong defaults is a question for follow-up studies.

\section{Limitations}
Each cell has three rounds on Magnific and six on Higgsfield, so per-model results are descriptive. We used three fixed target times, one style of reference and English prompts only. The models were reached through two commercial platforms: we cannot see the exact model versions, any internal prompt processing, or settings that the platforms do not expose. The readers are AI models; agreement is high and disagreements were arbitrated, but agreement is not accuracy: errors shared by both readers would pass unnoticed, especially on distorted dials, and a human audit of a random sample of agreed dials has not been done yet. The second Claude reader changed from Opus to Sonnet between study and replication, and three Sonnet readers reported measuring hands on the pixels in addition to looking. The replication repeats the same models on another platform and is not an independent sample of models; the share of images at 10:10 in condition A weights the current models more, as they have three rounds. In condition D the reference also changed the appearance of the dials in some models, so D does not isolate the geometric information. The origin of the bias in the training data is a plausible hypothesis that our design does not test. Finally, the platforms' terms forbid using the generated images to train models, which limits the licence of the released images.

\section*{Data availability}
All \nImages{} generated images, every prompt, the reference dials, the raw readings of every reader, the arbitration, the final class of every dial and a script that recomputes every number of this paper from the raw readings are available on Zenodo \citep{cazzaniga2026dataset} (\url{https://doi.org/10.5281/zenodo.23224681}). Images are released under a licence that allows any use, including commercial, but forbids training AI models on them; data and documentation under CC~BY~4.0; code under MIT.

\section*{Use of AI tools}
The images were generated by the AI models under study. The dials were read by AI models (OpenAI, Anthropic and Google) under the blind protocol described above, with human decisions on unresolved cases. The study design was reviewed with an OpenAI model before data collection. Scripts, analysis and the drafting of this paper were carried out with the assistance of Claude (Anthropic); the author reviewed all content and takes full responsibility for it.

\bibliographystyle{plainnat}
\bibliography{refs}

\appendix
\section{Prompts}
\label{app:prompts}
The four prompts for the wall clocks (S4), rotation 1. The other scenes change the subject, design, setting, camera, lighting and style lines; all prompts are in the dataset. In condition A the constraint line reads ``identical design for all three;'' without ``each one set as listed above''.

\begin{small}
\textbf{A, no time.}
\begin{verbatim}
Subject: three identical round analog wall clocks hanging in a single
  horizontal row on a plain wall
Clock design: thin black metal frame about 30 cm across, white dial, black
  Arabic numerals 1 to 12, two black hands only: a short thick hour hand
  reaching about half of the dial radius and a long thin minute hand
  reaching the minute track, both starting from the same small central
  pin; no second hand
Setting: smooth light grey painted wall, clean minimal interior
Composition: all three clocks fully visible, evenly spaced, same size,
  centered
Camera: straight-on frontal view, eye level, 50mm lens, sharp focus
  across all dials
Lighting: soft diffused daylight, even illumination, no glare on the glass
Style: photorealistic interior photography
Constraints: identical design for all three; Arabic numerals only, no
  other lettering, no logos, no brand names
Aspect ratio: 3:2
\end{verbatim}
\textbf{B, time in digits} adds after the subject line:
\begin{verbatim}
Times, left to right: 2:35, 6:50, 11:20
\end{verbatim}
\textbf{C, hands in words} adds instead:
\begin{verbatim}
Hands, left to right:
- clock 1: the hour hand points seven twelfths of the way from the 2 to
  the 3; the minute hand points exactly at the 7
- clock 2: the hour hand points five sixths of the way from the 6 to the
  7; the minute hand points exactly at the 10
- clock 3: the hour hand points one third of the way from the 11 to the
  12; the minute hand points exactly at the 4
\end{verbatim}
\textbf{D, words plus reference} adds to C, with the reference image attached:
\begin{verbatim}
Reference: use the attached diagram solely to determine the hand angles
  and their left-to-right assignment. Render the clocks using the design,
  scene and photographic style specified here.
\end{verbatim}
\end{small}

\section{The 32 older models, condition A}
\label{app:old}
Dials at 10:10 (out of 9) and images with all three clocks at 10:10 (out of 3), one round per scene.

\begin{table}[h]
\centering
\footnotesize
\begin{tabular}{lcc|lcc}
\toprule
Model & 10:10 dials & All three & Model & 10:10 dials & All three \\
\midrule
\csname @@input\endcsname tables/oldmodels.tex 
\bottomrule
\end{tabular}
\end{table}

\end{document}